\documentclass{article}
\PassOptionsToPackage{authoryear,round}{natbib}
\usepackage[dblblindworkshop,final]{neurips_2026}
\workshoptitle{The BabyVLM Workshop: Toward Developmentally Plausible Multimodal Systems}

\usepackage[utf8]{inputenc}
\usepackage[T1]{fontenc}
\usepackage{amsmath,amsfonts}
\usepackage{graphicx}
\usepackage{booktabs}
\usepackage{microtype}
\usepackage{xcolor}
\usepackage{xurl}
\usepackage[hidelinks]{hyperref}
\title{Probing Where Surprise Happens: Localized Violation-of-Expectation Evaluation of Video World Models}
\author{Yanbin Xu \\
Department of Psychology \\
University of Miami \\
Coral Gables, FL 33146 \\
\texttt{yxx744@miami.edu}
\And
Ming Bo Cai \\
Department of Psychology \\
University of Miami \\
Coral Gables, FL 33146\\
\texttt{mingbo.cai@miami.edu}}
\begin{document}
\maketitle
\begin{abstract}
Violation-of-expectation (VOE) benchmarks evaluate intuitive physics in video world models by testing whether physically impossible events produce greater prediction errors than matched possible events. However, conventional VOE evaluations aggregate prediction errors across space and time, making it unclear whether successful discrimination reflects model's understanding of physical violations at the locations and time where they occur. To address this limitation, we introduce localized VOE evaluation, a framework that measures prediction error differences between possible and impossible events within spatial regions and temporal intervals associated with physical violations. We compare global, temporally localized, and spatiotemporally localized evaluations of V-JEPA models pretrained on developmental egocentric videos or large-scale natural video datasets. Our results reveal that global and localized judgments can diverge in both directions: global aggregation can either obscure correct localized discrimination or produce apparent discrimination without corresponding localized evidence. During continued pretraining on developmental egocentric videos, agreement between global and localized judgments changes non-monotonically, while confidence in localized judgments increases despite modest improvements in accuracy. Together, these findings establish localized VOE evaluation as a complementary diagnostic for assessing intuitive physics in video world models and demonstrate that global prediction errors do not necessarily reflect the spatiotemporal evidence of physical violations.
\end{abstract}

\section{Introduction}

Intuitive physics, including the ability to understand, predict, and plan around physical objects, is fundamental to human intelligence. To interact and plan with the physical world like humans, it is suggested that embodied agents and world models should include similar internalized physical principles \citep{piloto2022,garrido2025,ha2018,su2025}.

A growing body of work evaluates physical understanding in video models using the \emph{violation-of-expectation} (VOE) paradigm, originating from developmental psychology \citep{riochet2022,piloto2022,garrido2025,bordes2025}. In a typical computational VOE evaluation, models are presented with matched physically possible and impossible scenes. Greater surprise, typically measured by L1 or L2 prediction error, for the impossible event is interpreted as evidence that the model has learned the relevant physical expectation. Using this approach, prior work has reported sensitivity to physical violations in V-JEPA models pretrained on large-scale natural videos \citep{garrido2025}, while evaluations of V-JEPA models have found more limited performance on benchmarks such as IntPhys 2 (hereafter IntPhys 2) \citep{su2025,bordes2025}.

However, existing VOE evaluation metrics aggregate prediction errors across spatial locations and temporal windows, typically reducing each video to a single mean or maximum score \citep{garrido2025,su2025}. Although these metrics assess whether a model distinguishes physically possible from impossible events, they leave an important question unresolved: when a world model successfully discriminates between possible and impossible events, is its prediction error concentrated at the locations and times where the physical violation occurs? This distinction is critical because a model with accurate physical expectations should exhibit prediction errors that are sensitive to the spatiotemporal evidence of a violation, rather than relying on signals from unrelated regions or intervals. Moreover, physical violations may be confined to small spatial regions or brief temporal intervals, allowing global aggregation to obscure meaningful localized prediction errors. Conversely, elevated prediction errors in irrelevant regions or intervals may produce apparent discrimination without reflecting sensitivity to the violation itself. Consequently, aggregate VOE performance alone cannot determine whether a model's predictions are aligned with the physical evidence underlying its judgments.

To address this limitation, we introduce \emph{localized violation-of-expectation} (localized VOE), a spatially and temporally grounded extension of predictive VOE evaluation. We annotate \emph{violation evidence regions and intervals} that specify where and when a physical violation becomes observable and measure possible--impossible prediction error differences specifically within those regions and intervals. This enables us to test whether the surprise expressed by a video world model corresponds to the physical event responsible for the violation. We apply this framework to evaluate V-JEPA models pretrained on developmental egocentric datasets, including BabyView \citep{long2024} and SAYCam \citep{sullivan2021}, as well as models pretrained on large scale natural video datasets. 

Our contributions are threefold:

\begin{enumerate}
\item \textbf{Localized VOE evaluation and annotations.} We introduce spatiotemporal violation evidence annotations for various VOE datasets that measure prediction error differences where and when physical violations become observable.

\item \textbf{Global-local diagnostic comparison.} We systematically compare global, temporally localized, and spatiotemporally localized VOE surprise scores across V-JEPA models, identifying when aggregate physical discrimination agrees with or diverges from event-localized sensitivity. We track global and localized VOE signals across BabyView pretraining to examine how global-local agreement, accurate judgment confidence, and VOE accuracy change through the exposure of developmental egocentric experience.

\end{enumerate}

\section{Related Work}

\paragraph{Computational accounts of intuitive physics and world models.}

A prominent computational account of human intuitive physics is \emph{probabilistic simulation}, which proposes that people reason about physical events by approximately simulating their future dynamics using an internal ``physics engine'' \citep{battaglia2013}. This view has motivated computational models that represent objects and their interactions explicitly. For example, ADEPT combines coarse object representations with probabilistic physical inference and evaluates model surprise against human judgments of physical events \citep{smith2019}. PLATO similarly learns object-centric representations and physical expectations from visual experience and introduces the Physical Concepts dataset for systematic evaluation \citep{piloto2022}.

Related ideas have increasingly been incorporated into world models, where physical regularities are suggested as the foundation for reasoning and prediction \citep{lecun2022,ha2018}. \citet{garrido2025}, for example, evaluate V-JEPA without task specific training and show that latent prediction error can discriminate between types of physically plausible and implausible events. \citet{su2025} examine whether developmentally inspired pretraining on SAYCam produces similar physical expectations, reporting limited performance on IntPhys 2. Together with our V-JEPA models trained on BabyView \citep{long2024}, these models provide an opportunity to compare how physical expectations emerge under different amounts and distributions of naturalistic visual experience. 

\paragraph{Violation-of-expectation benchmarks.}

The VOE paradigm, originating in developmental psychology, has become a common framework for evaluating whether computational models distinguish physically plausible from implausible events \citep{margoni2024,smith2019,riochet2022}. IntPhys 1, InfLevel, and GRASP evaluate complementary aspects of physical plausibility across controlled visual scenarios \citep{riochet2022,weihs2022,jassim2024}, while IntPhys 2 extends this setting to more visually complex environments with substantial camera and scene motion \citep{bordes2025}. Existing evaluations, however, typically aggregate model surprise across large spatial regions or temporal windows. Such aggregation makes it difficult to determine whether a model is surprised by the physical violation itself or by other differences elsewhere in the scene. Our work extends the VOE framework by annotating the spatiotemporal regions in which observable evidence of a violation occurs and evaluating prediction error specifically within these regions.

\paragraph{Localized surprise in human infants.}

Developmental psychology literature likewise suggests that aggregate measures of surprise can obscure when and where infants respond to unexpected events. The interpretation of infant VOE behavior depends critically on both experimental controls and the behavioral measures used to quantify surprise \citep{margoni2024}. \citet{yeung2016}, for example, show that region specific looking behavior can reveal sensitivity to unexpected events that is not apparent from total looking time alone. Similarly, in an object permanence paradigm, \citet{sirois2012} demonstrate that time resolved pupil responses reveal dynamics that would be obscured by temporally aggregated measures.

\section{Benchmark}

\subsection{Prediction Error and Paired VOE Scores}

In a machine learning VOE task, a model is presented with matched possible and impossible scenes that are identical except for a physical violation event. A model is considered sensitive to the tested physical principle if it assigns greater surprise to the impossible outcome. Given past video context, the model predicts the latent representation of future frames, and prediction error measures the discrepancy between predicted and ground truth representations. For each matched possible--impossible video pair, we define the paired VOE (surprise) score as the difference in mean L1 prediction error between the impossible and possible videos (Figure~\ref{fig:benchmark}):
\begin{equation}
S_{\mathrm{VOE}} = E_{\mathrm{impossible}} - E_{\mathrm{possible}}.
\label{eq:voe}
\end{equation}
A positive score therefore indicates greater prediction error for the physically impossible event and is counted as a correct VOE discrimination.

\subsection{Spatiotemporal Violation Evidence Annotations}

Standard VOE benchmarks generally identify which video contains a physical violation but do not explicitly specify when the violation becomes visually observable or where its evidence appears in the scene \citep{riochet2022,bordes2025,piloto2022}. To enable localized evaluation, we construct spatiotemporal violation evidence annotations for each matched possible-impossible pair.

\begin{figure}[tbp]
\centering
\includegraphics[width=\linewidth]{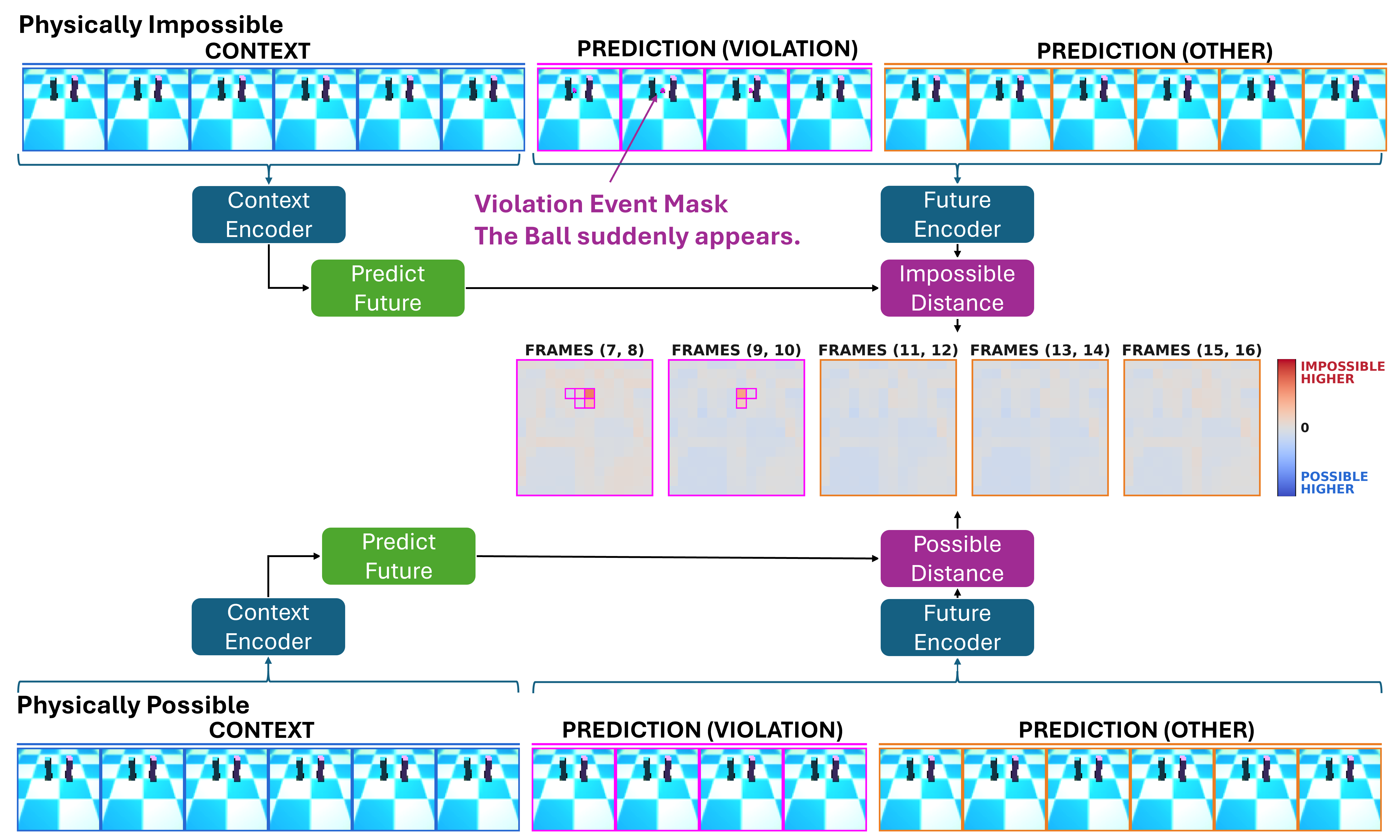}
\caption{\textbf{Localized VOE evaluation of V-JEPA.} Predictions from $C$ context frames are evaluated over $N$ future frames. Spatiotemporal violation evidence annotations identify event frames (magenta) and violation patches (magenta outlines). Local scores use annotated violation patches, whereas global scores average over all predicted tubelets (both temporally and spatially). Heatmaps show impossible-minus-possible prediction errors (Surprise score). Red indicates greater prediction error for the impossible video. The figure provides a simplified schematic for illustration. In the actual implementation, the future encoder processes both context and prediction frames. After that, the relevant future tokens are selected for comparison.}
\label{fig:benchmark}
\end{figure}

We define the \emph{violation onset} as the earliest frame at which the physically possible and impossible videos diverge in the visual state of the event relevant object (Figure~\ref{fig:benchmark}). When object masks are available, we identify this onset as the first frame at which the corresponding masks differ. For each subsequent frame,  the \emph{spatial violation region} is defined as the pixel-wise difference between the matched object masks. For IntPhys 2, where suitable object masks are not directly available, we generate them by segmenting the event relevant objects using the GazeBehavior Annotation Toolkit (GBAT) \citep{baig2026}.

We then map pixel level annotations onto the tokenization used by the video encoder. A visual patch is labeled as a \emph{violation patch} if it overlaps with the annotated violation region. For V-JEPA, each token corresponds to a temporal tubelet spanning two frames. We define the tubelet's spatial violation annotation as the union of the violation-patch annotations across those two frames. This produces token level annotations specifying both the temporal onset and spatial region of observable violation evidence.

\subsection{Global, Temporal, and Spatiotemporal Surprise}

We evaluate model surprise at three levels of scope. \emph{Global surprise} ($S_{\mathrm{glob}}$) averages prediction error over all predicted spatiotemporal tokens and therefore corresponds most closely to conventional VOE evaluation. \emph{Temporal surprise} ($S_{\mathrm{temp}}$) restricts evaluation to predictions within the temporal window of the violation event while averaging over the full spatial extent. \emph{Spatiotemporal surprise} ($S_{\mathrm{st}}$) further restricts prediction error to tokens that overlap with the annotated violation regions. For each level, the corresponding VOE score is obtained by subtracting the possible video prediction error from the impossible video prediction error. Together, these three measurements test not only whether a model assigns greater overall surprise to impossible events, but also whether that surprise emerges at the time and location where evidence for the physical violation becomes observable.

For Physical Concepts, we extend each 15-frame video to 16 frames by repeating the final frame, allowing V-JEPA to represent the sequence as eight two-frame tubelets. Frames preceding the violation onset are used as the prediction context. Frames after the violation onset serve as the prediction target. Global surprise is evaluated over all spatial tokens after the onset, whereas spatiotemporal surprise is evaluated only over annotated violation tokens.

For IntPhys 1, we temporally subsample each 100-frame video by retaining every second frame, resulting in 50 frames. For global surprise, we apply a sliding 16 frame window with a stride of two subsampled frames. Within each window, the first $C$ frames serve as context, and the remaining $16-C$ frames are prediction targets, where $C$ is an evaluation hyperparameter. We compute the mean prediction error across all prediction target tokens within each window and then average across windows to obtain a single global surprise score. For temporally localized surprise, we restrict prediction error evaluation to all spatial tokens within the violation event interval, without using sliding windows. For spatiotemporally localized surprise, we further restrict evaluation to tokens within the spatial violation region, such that only prediction errors corresponding to both the violation interval and the relevant spatial locations contribute to the surprise score.

\section{Experimental Setup}

\subsection{Models and Pretraining}

We pretrained a V-JEPA base model with $16 \times 16$ spatial patches and an input resolution of $224 \times 224$ (V-JEPA-B-BabyView), following the self-supervised feature prediction procedure introduced by \citet{bardes2024}. The encoder contains 86.2 million parameters, and the predictor adds 21.9 million trainable parameters. BabyView comprises naturalistic video recorded using head mounted cameras worn by infants and young children aged 6--36 months; our training recordings were drawn from the approximately 868-hour corpus \citep{long2024}. The final BabyView checkpoint completed 83,250 optimization steps with a global batch size of 32, corresponding to 2.664 million clip presentations, or approximately 1,500 hours of exposure when each clip is treated as roughly two seconds long.

We compared this model with three publicly released V-JEPA checkpoints differing in model scale and pretraining data. V-JEPA-B-SAYCam \citep{su2025} uses a similarly sized backbone trained on SAYCam, a longitudinal egocentric dataset containing approximately 472 hours of recordings from three children aged 6--32 months \citep{sullivan2021}. We evaluated the released checkpoint without additional training. Its stored optimizer state indicates 14,601 updates with a global batch size of 32, corresponding to 467,232 clip presentations, or approximately 260 hours of cumulative exposure. V-JEPA-1-H-VM2M is a ViT-H/16 model with approximately 632 million encoder parameters and 655 million parameters including its predictor. It was pretrained on VideoMix2M for 90,000 iterations with a batch size of 3,072, corresponding to approximately 276.5 million clip presentations. The model is trained on VideoMix2M, which combines approximately two million videos from Something-Something v2, Kinetics, and HowTo100M \citep{bardes2024}. V-JEPA-2-H-VM22M is a ViT-H/16 model, with the same architecture as V-JEPA-1-H-VM2M. This public checkpoint records 132,000 optimizer updates. With a global batch size of 3,072, this corresponds to approximately 405.5 million training sample presentations. The model is trained on VideoMix22M, which contains approximately 22 million samples.\citep{assran2025}.

\section{Results}

\subsection{Prediction error differences (surprise score) need not favor the impossible event}

\begin{figure}[tbp]
\centering
\includegraphics[width=\linewidth]{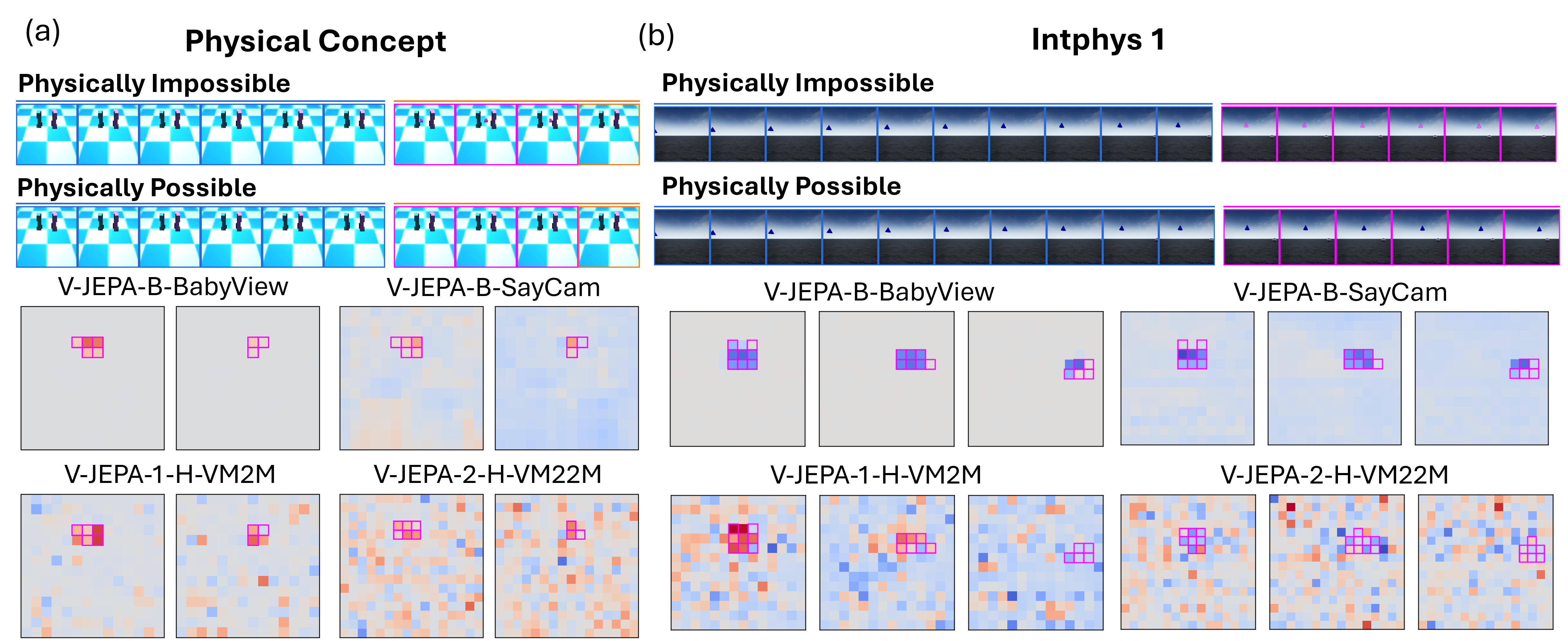}
\caption{\textbf{V-JEPA models show local surprise scores.} Examples of spatiotemporally localized surprise scores for matched physically possible and impossible video pairs from the Physical Concepts and IntPhys 1 datasets. Frame labels follow the same convention as in Figure 1. Each heatmap shows the surprise score map for the corresponding video pair produced by a different model, with patches containing the physical violation outlined in magenta. Red indicates greater prediction error for the impossible video. Because two consecutive frames are grouped into a single tubelet, each heatmap corresponds to a pair of consecutive frames. These two examples, one from each dataset, are illustrative and should not be interpreted as evidence that IntPhys 1 is generally more challenging than the Physical Concepts dataset.}
\label{fig:local_maps}
\end{figure}

Figure~\ref{fig:local_maps} visualizes token level prediction error differences for matched possible-impossible video pairs (surprise score). For an example pair from the Physical Concepts dataset (Figure~\ref{fig:local_maps}a), positive differences overlap with annotated violation patches (indicated by magenta boxes), showing that prediction error differences are elevated at the location of the violation. This pattern is visible in models trained in both developmental egocentric and large-scale natural-video pretraining settings.

Another example from IntPhys 1 (Figure~\ref{fig:local_maps}b) shows an opposite pattern that negative surprise scores are localized at the violation region. These negative responses indicate three models (BabyView-trained, SAYCam-trained, and V-JEPA-2-H-VM22M \citep{assran2025}) show greater prediction error for the possible event. We postulate that these negative prediction errors might be induced by the motion of objects in the physically possible video which is absent in the impossible video, but this interpretation needs further testing. Prediction error differences also extend beyond the annotated regions, particularly in the more extensively trained V-JEPA-1-H-VM2M and V-JEPA-2-H-VM22M maps. These examples shown here motivate us to further explore the difference of physical plausibility discrimination judgments between localized and global VOE surprise signals.

\subsection{Global and localized evaluation can yield distinct conclusions on physical plausibility judgment}

\begin{figure}[tbp]
\centering
\includegraphics[width=\linewidth]{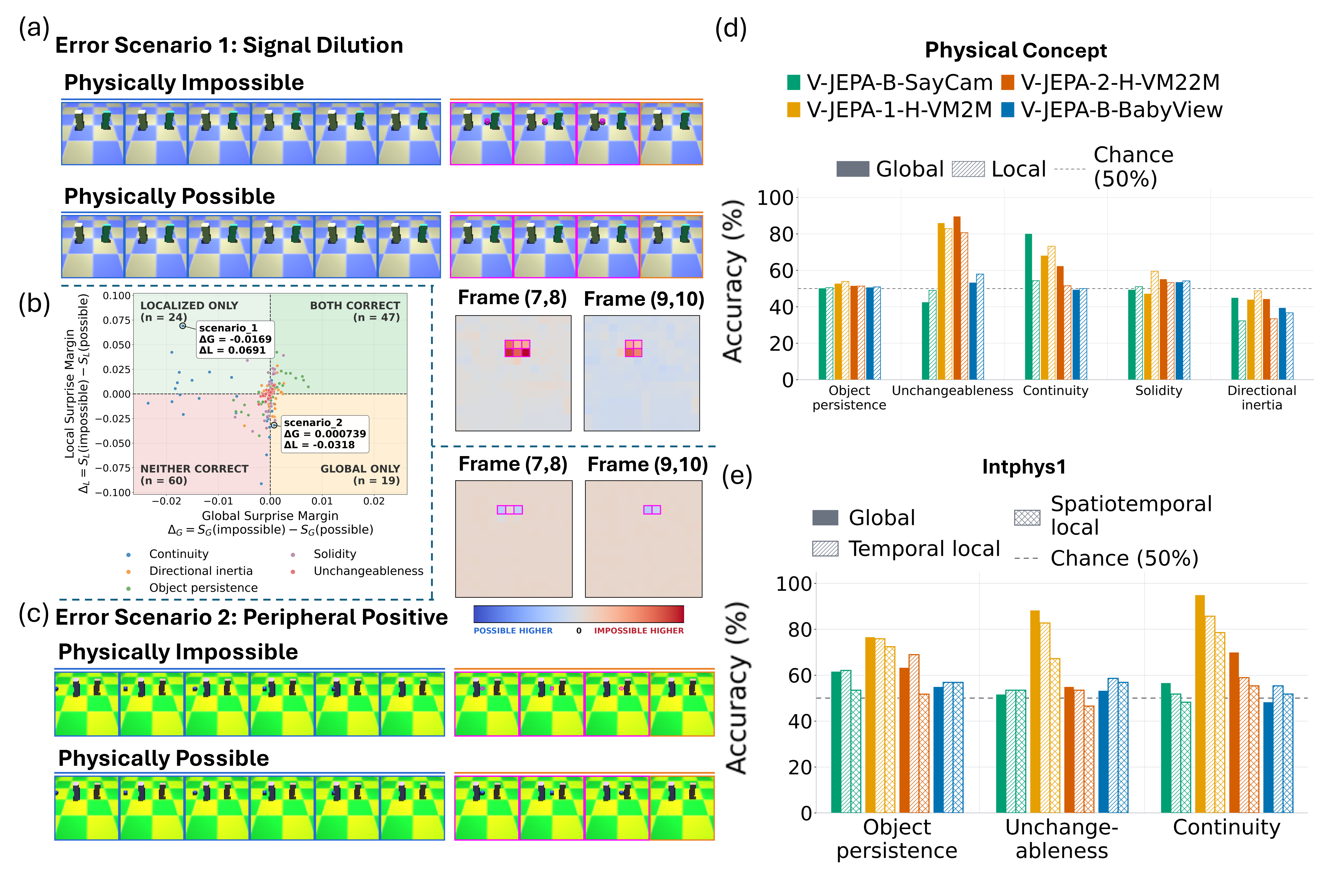}
\caption{\textbf{Global and localized evaluations can yield different judgments of physical plausibility.} \textbf{(a--c)} Two example pairs from the Physical Concepts dataset illustrate judgment disagreement. The scatter plot compares global (x-axis) and localized (y-axis) surprise scores, defined as prediction error for the impossible video minus that for the possible video. Positive scores indicate correct discrimination. The four quadrants distinguish pairs classified correctly by both evaluations, either evaluation alone, or neither. Heatmaps show the surprise score, with red indicating greater error for the impossible video and blue indicating the opposite. Magenta outlines mark annotated violation regions. In Scenario 1, spatial averaging obscures the localized violation signal, yielding correct localized but incorrect global discrimination. In Scenario 2, elevated background error produces a positive global margin despite a negative localized margin. \textbf{(d)} Global and localized accuracy across four V-JEPA models and five physical principles in Physical Concepts. \textbf{(e)} Global, spatiotemporal-local, and spatial-local accuracy across the same models and three physical principles in IntPhys 1. Dashed lines indicate chance performance (50\%).}
\label{fig:global_local}
\end{figure}

Global and localized VOE scores can yield conflicting judgments for the same video pair (Figure~\ref{fig:global_local}(a--c)). To illustrate this discrepancy, we compare global and localized surprise score using two example pairs from the Physical Concepts Dataset, evaluated with V-JEPA-B-BabyView-step-10000. We identify two scenarios. In Scenario 1, the localized surprise score is positive (0.0691), whereas the global surprise score is negative ($-0.0169$). This indicates that averaging prediction errors across all predicted tubelets, both spatially and temporally, can reverse the judgment derived from the annotated violation region. In Scenario 2, the opposite pattern emerges: the global surprise score is positive (0.000739), whereas the localized score is incorrectly negative ($-0.0318$). In this case, positive surprise score outside the annotated violation region hides the misleading negative score within it, even though it gives a correct global VOE judgment. Together, these examples demonstrate how global aggregation can obscure or misrepresent model responses to localized physical violations.

We then systematically evaluate the global-local agreement across all pairs of Physical Concepts for V-JEPA-B-BabyView-step-10000 (Figure~\ref{fig:global_local}(b)). We found that 24 pairs classified correctly only by localized evaluation and 19 only by global evaluation, compared with 47 classified correctly by both and 60 by neither. The evaluations therefore disagree on 28.7\% pairs in Physical Concept.

We further evaluated if such differences in local and global surprise score eventually lead to accuracy differences (Figure~\ref{fig:global_local}(d--e)) on both IntPhys 1 and Physical Concepts dataset. We found that localization changes the diagnostic profile but is not universally advantageous. For example, for continuity in Physical Concepts, SAYCam accuracy falls from approximately 80\% globally to 54\% locally, whereas V-JEPA-1-H-VM2M increases from approximately 68\% to 73\%. For unchangeableness, both large-scale models retain relatively high accuracy under localization, while the accuracy for directional inertia remains below 50\% for all four models under both evaluations. 

\subsection{Localized surprise strengthens through BabyView pretraining, with modest, dataset-dependent accuracy gains}

\begin{figure}[tbp]
\centering
\includegraphics[width=\linewidth]{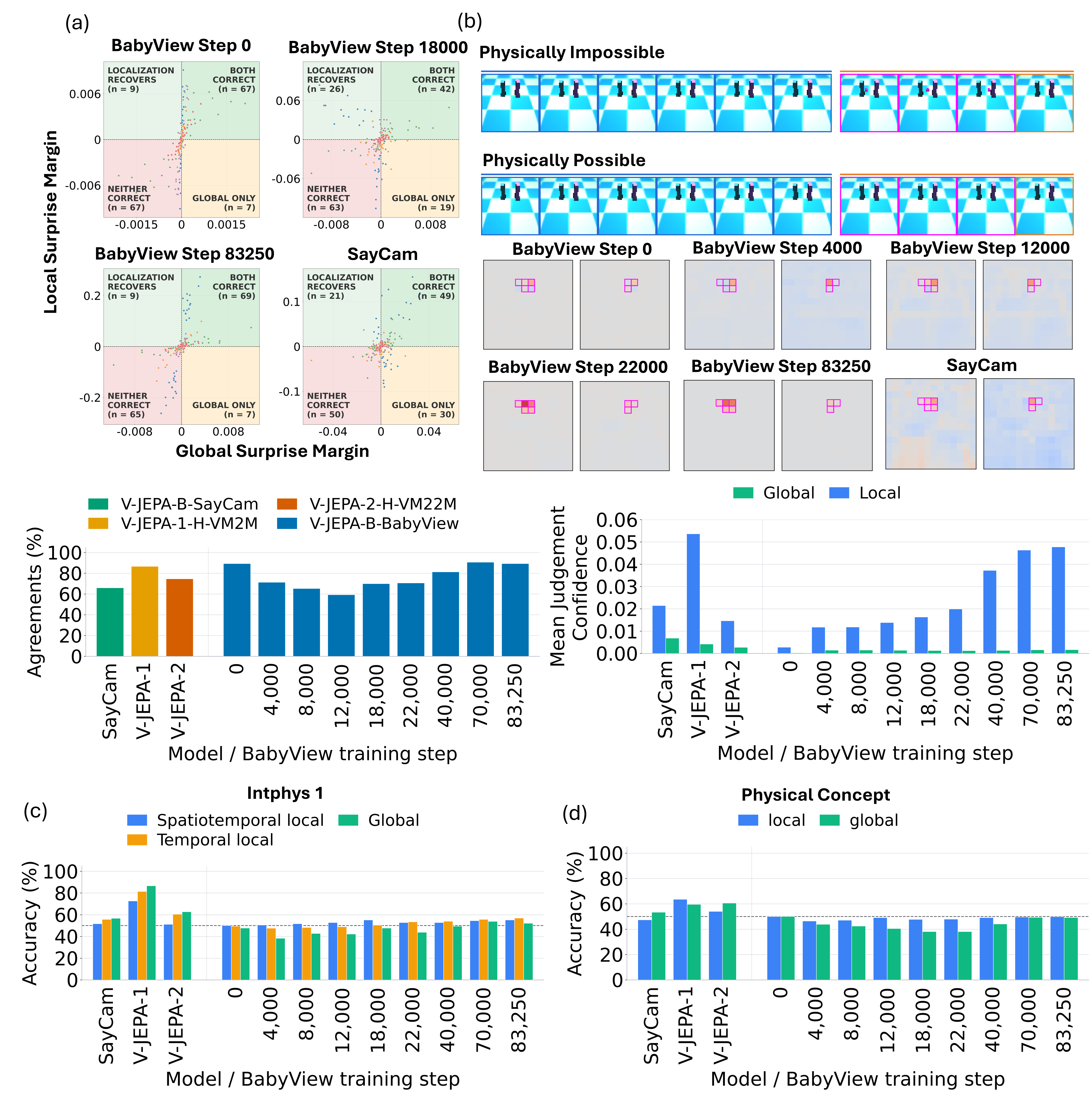}
\caption{\textbf{Global-local agreement, judgment confidence, and VOE accuracy across models and BabyView pretraining checkpoints.} \textbf{(a)} Global-local agreement is the percentage of Physical Concepts video pairs for which global and localized evaluations agree: both correctly assign higher surprise score to the impossible video (upper-right quadrant), or both fail to do so (lower-left quadrant). The bar chart summarizes agreement across all pairs for each model and checkpoint. Example scatterplots show global versus local surprise scores for individual pairs, colored by physical principles. \textbf{(b)} Example surprise score heatmaps with labeled violation patches are provided to visualize the change in prediction across pretraining. The bar chart summarizes mean surprise scores over pairs classified correctly by both evaluations for models and checkpoints. \textbf{(c--d)} Pairwise accuracy on Intphys 1  (c) and Physical Concepts (d) for the same models and checkpoints.}
\label{fig:learning}
\end{figure}

We next examine how global-local agreement, judgment confidence, and VOE accuracy change across BabyView pretraining checkpoints (Figure~\ref{fig:learning}). We define agreement as the percentage of video pairs for which global and localized evaluations yield the same judgment, whether correct or incorrect. On Physical Concepts, agreement follows a non-monotonic trajectory (Figure~\ref{fig:learning}(a)): it is approximately 89\% at initialization, declines to approximately 59\% at step 12,000, and recovers to approximately 89\% at step 83,250. We define judgment confidence as the mean paired surprise score, calculated separately for global and localized evaluations over pairs that both evaluations classify correctly. Localized judgment confidence rises through training, whereas global judgment confidence remains low with comparatively little changes (bottom part of Figure~\ref{fig:learning}(b)). These findings can also be visualized by the example (top part of Figure~\ref{fig:learning}(b)), where heatmap intensity increases within violation patches through training, and intensity within the peripheral patches first decreases to negative values and then increases back to 0 through training.

Changes in VOE accuracy are comparatively modest. The initialization and final checkpoint scatterplots both contain substantial groups of pairs for which neither evaluation is correct. Consistent with this observation, the Physical Concepts accuracy bars remain near 50\% at the final checkpoint (Figure~\ref{fig:learning}(c)). On IntPhys 1, the localized and temporal-window accuracy bars show modest gains from approximately 50\% at initialization to the mid-50\% range at the final checkpoint (Figure~\ref{fig:learning}(d)). These changes within a model are small relative to the differences between models (e.g., between BabyView and V-JEPA-1-H-VM2M). 
% Together, the checkpoint analyses explore changes in agreement, confidence, and discrimination: pretraining alters all, with increasing agreement and confidence, but there is not sufficient evidence of improved physical understanding.

\section{Discussion and Conclusion}

We introduced localized violation-of-expectation (VOE) evaluation to quantify model surprise at annotated violation locations and times. Our results reveal that global and localized judgments can disagree in both directions: global aggregation may dilute violation specific signals, producing false negatives, or amplify irrelevant prediction errors outside the violation region, producing disguising VOE signal. We further found that global-local agreement change non-monotonically, and localized judgment confidence increase over pretraining, despite limited improvements in discrimination accuracy. These findings highlight the importance of jointly reporting global and localized VOE performance.

\begin{ack}
YX is supported by the Antonio Orlando Award of the University of Miami.
\end{ack}

\clearpage
\bibliographystyle{plainnat}
\bibliography{references}
\clearpage

\appendix
\section{Appendix}
\subsection{BabyView Pretraining Details}

Our pretraining implementation closely followed the code and configurations introduced by \citet{bardes2024} and the subsequent SAYCam adaptation \citep{su2025}. We used a ViT-B/16 encoder and a 12-layer transformer predictor with a 384-dimensional predictor embedding. Each training example consisted of 16 frames sampled from a randomly selected 64-frame temporal span and resized to $224 \times 224$ pixels. A tubelet size of two frames produced eight temporal positions and a $14 \times 14$ spatial token grid. We applied a temporally consistent random resized crop, horizontal flipping, and ImageNet normalization. The predictor was trained with an L1 objective to match layer-normalized representations produced by an exponential-moving-average target encoder.

We partitioned the eligible recordings into participant-disjoint training and validation sets using a fixed random seed of 234. Recordings with fewer than 64 decodable frames were excluded, resulting in 7,660 training videos, 900 validation videos, and 16 excluded recordings. Training used a global batch size of 32, distributed as eight clips per GPU across four NVIDIA L40S GPUs. We trained for 83,250 optimizer updates, represented internally as 278 scheduling epochs of 300 updates each.

We optimized the encoder and predictor using AdamW with $\beta_1=0.9$, $\beta_2=0.999$, and $\epsilon=10^{-8}$. The learning rate increased from $2\times10^{-4}$ to $6.25\times10^{-4}$ during an 11,100-step warm-up and then followed a cosine schedule to $10^{-6}$. Weight decay was fixed at 0.04, gradients were clipped to a maximum norm of 1.0 after warm-up, and the target-encoder momentum increased linearly from 0.998 to 1.0. Training used bfloat16 autocasting and scaled dot-product attention. We saved checkpoints throughout training, including the initialization checkpoint and the final checkpoint at step 83,250; no early-stopping criterion was applied.

\newpage

\end{document}